\documentclass{./interspeech2025kit/Interspeech}

\usepackage{comment}
\usepackage{caption}
\usepackage[table]{xcolor}
\usepackage{subcaption}
\usepackage{array}

\usepackage{xcolor}

\usepackage{listings}
\definecolor{firebrick}{rgb}{0.7, 0.13, 0.13}

\usepackage{xspace}
\newcommand{\eng}{\texttt{eng}\xspace}
\newcommand{\spa}{\texttt{spa}\xspace}
\newcommand{\zho}{\texttt{zho}\xspace}

 \interspeechcameraready 

\newcommand\blfootnote[1]{%
  \begingroup
  \renewcommand\thefootnote{}\footnote{#1}%
  \addtocounter{footnote}{-1}%
  \endgroup
}

\title{Effects of Speaker Count, Duration, and Accent Diversity on Zero-Shot Accent Robustness in Low-Resource ASR}

\author[affiliation={1,*}]{Zheng-Xin}{Yong}
\author[affiliation={2}]{Vineel}{Pratap}
\author[affiliation={}]{Michael}{Auli$^\dagger$}
\author[affiliation={2}]{Jean}{Maillard}

\affiliation{Department of Computer Science}{Brown University}{United States}
\affiliation{}{Meta FAIR}{United States}
\email{contact.yong@brown.edu, jeanm@meta.com}

\begin{document}

\maketitle

\begin{abstract}
    
To build an automatic speech recognition (ASR) system that can serve everyone in the world, the ASR needs to be robust to a wide range of accents including unseen accents. We systematically study how three different variables in training data---the number of speakers, the audio duration per each individual speaker, and the diversity of accents---affect ASR robustness towards unseen accents in a low-resource training regime. We observe that for a fixed number of ASR training hours, it is more beneficial to increase the number of speakers (which means each speaker contributes less) than the number of hours contributed per speaker. We also observe that more speakers enables ASR performance gains from scaling number of hours. Surprisingly, we observe minimal benefits to prioritizing speakers with different accents when the number of speakers is controlled. Our work suggests that practitioners should prioritize increasing the speaker count in ASR training data composition for new languages.\blfootnote{\footnotesize $^*$Work done during internship at Meta.}\blfootnote{\footnotesize $^\dagger$Work done while at Meta.}
\end{abstract}

\section{Introduction}

Automatic speech recognition (ASR) systems have become an integral part of our daily lives, powering virtual assistants, transcription services, and accessibility tools \cite{prabhavalkar2023end}. However, these systems often exhibit significant performance disparities across different accents, potentially excluding or poorly serving large segments of the global population \cite{koenecke2020racial,
prinos2024speaking}. While recent advances have improved overall ASR accuracy, the challenge of accent robustness remains a critical barrier to developing truly inclusive speech technology.

To address this challenge, we need a systematic understanding of how different aspects of training data collection and composition affect ASR systems' ability to handle accent variation, particularly for accents not represented in the training data. 
Specifically, our research question is as follows: \textbf{how does scaling each of the three dimensions---number of speakers, number of hours per speaker, and accent diversity---in training data in a low-resource setting affect ASR performance on accents \emph{outside} the training distribution?} This is also known as \textit{zero-shot accent robustness} evaluation \cite{hinsvark2021accented}. We experiment with three languages that have high-quality datasets with accent information readily labeled, namely English, Spanish and Mandarin Chinese, and our work covers both L1/L2 accents (for English) as well as regional accents (for Spanish and Mandarin Chinese).

Our results offer several novel actionable insights for building ASR systems robust to out-of-distribution accents. First, for a fixed budget of training hours, models trained with more speakers (but less data per speaker) consistently outperform those trained with fewer speakers (but more data per speaker). Furthermore, having more speakers enables better utilization of additional training hours. Surprisingly, in low-resource settings, increasing accent diversity in training data yields minimal benefits when controlling for the total number of speakers and hours. This suggests that accent coverage may be less critical than previously thought for zero-shot accent generalization.

\begin{table*}[!t]
\small
\centering

\caption{Information about accent distribution for our zero-shot accent robustness study. We \textbf{bold} the accents used when we vary the number of speakers (Section~\ref{subsec:training-num-speakers}) and speaker duration (Section~\ref{subsec:training-speaking-duration}). These accents are the dominant accent, whereas the accents in parentheses (\texttt{+ ...}) are the additional accents when we vary accent diversity during training (Section~\ref{subsec:training-accent-diversity}).}
\label{tab:accent-split-summary}
\begin{tabular}{ll p{0.3\linewidth} p{0.3\linewidth}}
\toprule
Languages & Accent Split & Seen Accents (Training/Validation) & Out-Of-Distribution Accents (Test) \\

\midrule
English & L1/L2 & \textbf{American}, (\texttt{+} Irish, British, Canadian, New Zealand) & Hindi, Korean, Mandarin, Spanish, Arabic and Vietnamese (6) \\

Spanish & regional & \textbf{Mexico}, (\texttt{+} Central America, Caribe, Rioplatense, Andean-Pacific) & South-center Spain, Canary Islands, South-peninsular Spain, North-peninsular Spain (4) \\

Mandarin Chinese & regional & \textbf{An Hui}, (\texttt{+} Ning Xia, Shang Hai, Guang Xi, Guang Dong) & Chong Qing, Gan Su, He Nan, Jiang Su, Jiang Xi, Shan Xi, Si Chuan (7) \\ 

\bottomrule
\end{tabular}
\end{table*}

\section{Related Work}
ASR systems have been found to exhibit bias towards certain accents. For instance, prior work discovered substantial word error rate (WER) differences between transcribing L1 (first language speaker) and L2 (second language speaker) English speech \cite{chan2022training,
dichristofano2024performance}. 
Regional accent biases, where ASR performance for certain regional dialects or accents is significantly better, are also observed for certain L1 varieties of English \cite{sanabria2023edinburgh} and other languages such as Brazilian Portuguese \cite{kulkarni-etal-2024-balancing}, Dutch \cite{zhang2022mitigating}, Modern Standard Arabic \cite{sawalha2013effects}, and Mandarin Chinese \cite{feng2024inclusiveasr}.

Prior work on reducing accent bias primarily focuses on improving ASR training by augmenting existing data, such as data augmentation via perturbation \cite{zhang2022comparing,zhang2023exploring} and voice-cloning \cite{baas2022lowresource,klumpp2023synthetic}, or through novel ASR training methods such as domain-adversarial training \cite{sun2018domain, 
zhang2022mitigating}, learnable accent-specific codebooks \cite{prabhu-etal-2023-accented},  and multi-task learning \cite{ghorbani18_interspeech, 
viglino2019end}. Nonetheless, little attention has been given to investigate how the training data composition affects bias against out-of-distribution accents \textit{in the first place}. 

One relevant work in this direction studies how different data partitions affect transfer learning of L1 English accent to L2 accents (and vice versa) \cite{shibano-etal-2021-speech}, but critical factors, such as number of speakers and the training hour contribution per each speaker, were \textit{not controlled} in their setup. 
Another relevant work approaches ASR by studying the effects of number of hours and speakers in pretraining and ASR training data \cite{dan2023morespeaking} but this prior work focuses only on the general English ASR performance. In contrast, our work focuses on transfer learning for accents, and we further include in our study an additional factor of accent variety during training.

\section{Experimental Setup}
Our work explores how (1) number of speakers, (2) audio duration per speaker and (3) explicit accent diversity in training data affect ASR performance on \textit{accents outside of the training distribution}. Primarily, our work focuses on the low-resource training regime, as our goal is to provide recommendations to practitioners about which of the three variables to prioritize during ASR data collection for new languages.

For English (\eng), we investigate non-native accent robustness, by training our models on L1 accents and evaluating on L2 accents in a zero-shot fashion. For Simplified Mandarin Chinese (\zho) and Spanish (\spa), we study regional accent robustness by splitting our training data based on regional accent information and evaluating on out-of-distribution regional accents. Table~\ref{tab:accent-split-summary} shows the split of accents.

\subsection{Training Data and Setup}
In the following, we describe three different experimental setups, each exploring how varying speaker count (Section~\ref{subsec:training-num-speakers}), speaker duration (Section~\ref{subsec:training-speaking-duration}), and accent diversity (Section~\ref{subsec:training-accent-diversity}), affects ASR zero-shot accent robustness.

\subsubsection{Number of Speakers}
\label{subsec:training-num-speakers}

For \eng, we first identify American speakers in the English training split of Multilingual LibriSpeech \cite{pratap2020mls} using LibriVox's Accent Table \cite{librivox2024accenttable}. 
For \spa, we obtain recordings by speakers from Mexico from the training split of the Common Voice Corpus 19.0 \cite{ardila-etal-2020-common} using the provided metadata. For \zho, we use MAGICDATA \cite{magicdata2019}, which consists of thousands of speakers from different accent areas in China, and identify speakers originating from the An Hui province. These are the accents learned during training.

For a fixed total hours of audio duration $T$, we vary the number of speakers $N$ to compose the training data.\footnote{We randomly select from a set of speakers in which each speaks at least $\frac{T}{N}$ hours.} For \spa and \zho, we experiment with up to $N=25$ different speakers, whereas for \eng we use up to $N=15$ different speakers. We repeat the same experiments on different $T$: for \eng and \spa, we use $T=\{5, 10, 15\}$ whereas for \zho,  the dataset characteristics limit our evaluation to $T=\{5, 11.25\}$. 

\subsubsection{Number of Training Hours Per Speaker}
\label{subsec:training-speaking-duration}

We use the same training data sources following Section~\ref{subsec:training-num-speakers}. Here, we fix $N$ and then vary the per-speaker speaking duration $t$ instead. All $N$ speakers contribute the exact same $t$ duration of audios in the training data; therefore, the total training audio duration $T$ is $N \times t$. For \eng and \spa, we vary $t$ up to 60 minutes; for \zho, we use up to $t=45$ minutes (due to constraints in the original data). For all three languages, we control $N=\{5, 10, 15\}$ randomly selected speakers.

We further experiment with \textbf{single-speaker training setup}, where we only use a single speaker and scale their per-speaker duration up to 40 hours. Here, the speaker's speaking duration makes up the entire audio training hours. We only experiment with \eng as \eng training dataset contains eight American speakers each of whom spoke more 40 hours.

\subsubsection{Accent Diversity}
\label{subsec:training-accent-diversity}
The ASR training data distribution is generally dominated by one (or very few) accent(s). For instance, in the GLOBE dataset \cite{wang2024globe} that provides accent label information for audio samples in the English Common Voice corpus \cite{ardila-etal-2020-common}, nearly 70\% of the training data are labeled as ``United States English'', closely followed by around 25\% of ``England English''. These two accents alone already make up nearly 95\% of the English training data. Therefore, our experiments for studying the effects of accent diversity mimics this uneven distribution–––it contains a \textit{dominant accent} where majority speakers share a single accent.

We fix both $N$ speakers and their per-speaker duration $t$, and we vary the total number of unique accents $K$ in the training data. Note that for the setups in Section~\ref{subsec:training-num-speakers} and Section~\ref{subsec:training-speaking-duration} we have $K=1$, because all speakers share the same accent. We increment $K$ by first randomly choosing a unique additional accent from Table~\ref{tab:accent-split-summary} and then randomly sampling one speaker with the accent to replace one of the dominant accent speakers. In other words, when $K>1$, we would have $K-1$ speakers of unique additional accents. For instance, if we set $K=3$ and $N=20$ for English training data, it means that we have 18 American speakers and 2 speakers of different accents (e.g., Irish and England accents).

For all languages, we vary up to $K=5$ different accents in training. For \eng, we have $N=20$ speakers each speaking for $t=60$ minutes (totaling 20 hours of ASR training data). For \spa and \zho, due to data constraint, we use ($N=15$ speakers, $t=60$ minutes) and ($N=15$ speakers, $t=45$ minutes) respectively.

\begin{figure*}[!t]
    \centering
    \includegraphics[width=0.95\linewidth]{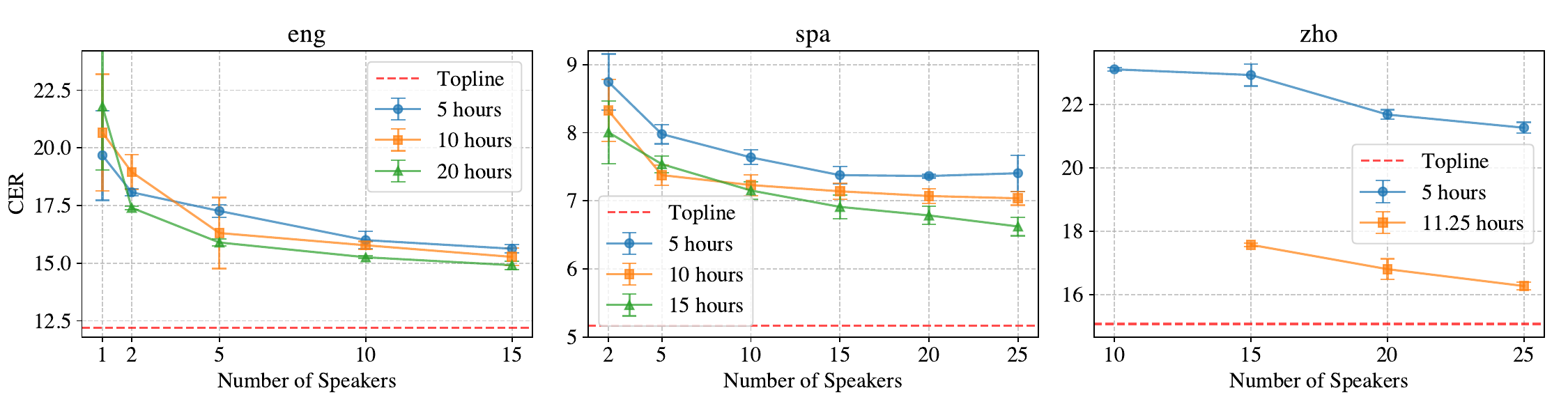}
    \caption{Effects of increasing the number of speakers on out-of-distribution accent ASR performance measured by CER (lower CER means better ASR). For each line, while we increase number of speakers, we kept the training data at a fixed total duration (hours) as shown in the legends.}
    \label{fig:more-speakers}
\end{figure*}

\begin{table*}[!t]
\centering
\small
\caption{
Effects of number of speakers and speaking duration on CER of out-of-distribution accents while keeping the total duration of the training data ($\text{\# speakers} \times \text{speaker duration}$) constant. Color shading is used to highlight that we should prioritize increasing speaker count over speaker duration to improve ASR zero-shot accent robustness.}
\label{tab:speakers-vs-speaking}
\begin{tabular}{lccccc}
\hline
Languages & Total Duration (min) & \# Speakers & Speaker Duration (min) & CER ($\downarrow$) \\
\hline
English (\eng) & 300 & \cellcolor{blue!1} 5 & 60 & \cellcolor{green!1} 17.3 {\scriptsize $\pm$ 0.3} \\
    & 300 & \cellcolor{blue!5} 10 & 30 & \cellcolor{green!10} 15.9 {\scriptsize $\pm$ 0.1} \\
    & 300 & \cellcolor{blue!10} 15 & 20 & \cellcolor{green!20} 15.7 {\scriptsize $\pm$ 0.1} \\
\hline
Spanish (\spa) & 300 & \cellcolor{blue!1} 5 & 60 & \cellcolor{green!1} 7.8 {\scriptsize $\pm$ 0.4} \\
    & 300 & \cellcolor{blue!5} 10 & 30 & \cellcolor{green!10} 7.5 {\scriptsize $\pm$ 0.1} \\
    & 300 & \cellcolor{blue!10} 15 & 20 & \cellcolor{green!20} 7.2 {\scriptsize $\pm$ 0.1} \\
\hline
Mandarin Chinese (\zho) & 150 & \cellcolor{blue!1} 5 & 30 & \cellcolor{green!1} 60.3 {\scriptsize $\pm$ 5.8} \\
    & 150 & \cellcolor{blue!5} 10 & 15 & \cellcolor{green!10} 51.6 {\scriptsize $\pm$ 3.2} \\
    & 150 & \cellcolor{blue!10} 15 & 10 & \cellcolor{green!20} 39.9 {\scriptsize $\pm$ 1.9} \\
\hline
\end{tabular}
\end{table*}

\subsubsection{Model Training}\label{sec:training-setup}
Since we work on accents in multilingual settings, we use the multilingual pretrained MMS model with 300 million parameters \cite{pratap2024scaling} as our base model for ASR training. 
We follow the same ASR training code and hyperparameter configurations open-sourced by \cite{pratap2024scaling}. 
We construct a one-hour validation dataset to choose the best checkpoint for ASR, and we train all ASR models to convergence. For result robustness, we train three different models using different random subsets of speakers wherever possible; otherwise (in scenarios where all possible speakers are already included), we train with different seeds.

\subsection{Evaluation Setup}
We evaluate zero-shot accent robustness using L2-ARCTIC datasets \cite{zhao2018l2arctic} for \eng, MAGICDATA \cite{magicdata2019} for \zho, and CommonVoice Corpus 19.0 \cite{ardila-etal-2020-common} for \spa. For \eng, L2-ARCTIC contain 24 non-native speakers of English with a total of six different accents, and we describe the statistics for L2-ARCTIC in the following paragraph that characterizes toplines. Our \zho test data consist of 43 speakers and 28.1 hours of total audio duration, whereas our \spa consists of 647 speakers and 2.3 hours of data. For \zho and \spa, we select the test data according to Table~\ref{tab:accent-split-summary} using the annotated regional accent labels in the test splits of MAGICDATA and CommonVoice.

In our work, we report the \textit{topline} results, where the ASR models are trained on speakers with test accents. In other words, toplines are trained with the best possible data as the test accents are in-distribution. For \eng, since L2-ARCTIC has four speakers for each non-native accents, we randomly select one speaker from each non-native accents to build our \eng test data. This test set consists of around 3.2 hours of audio data in total. The rest of the L2-ARCTIC data are used to train the English topline model. For \zho and \spa, we take 11.25 hours and 15 hours of data labeled with test accents from the train splits of MAGICDATA and CommonVoice respectively.

In addition, we evaluate the \eng single-speaker setup (described in Section~\ref{subsec:training-speaking-duration}) on the entire L1-ARCTIC dataset \cite{kominek2004l1arctic}, which comprises around 4500 utterances from L1 American English speakers, and the entire \eng test split of FLEURS \cite{conneau2023fleurs}. We refer readers to the cited sources for data distribution details.

We report the Character Error Rate (CER) for all of our experiments.\footnote{We observe strong correlation between CER and Word Error Rate (WER) ($r=0.92$) in our evaluation, so we report CER only for brevity.} The lower the CER, the better the ASR performance. We also report standard deviations from  different training runs as mentioned in Section~\ref{sec:training-setup}.

\begin{figure}[!t]
    \centering
    \includegraphics[width=0.99\linewidth]{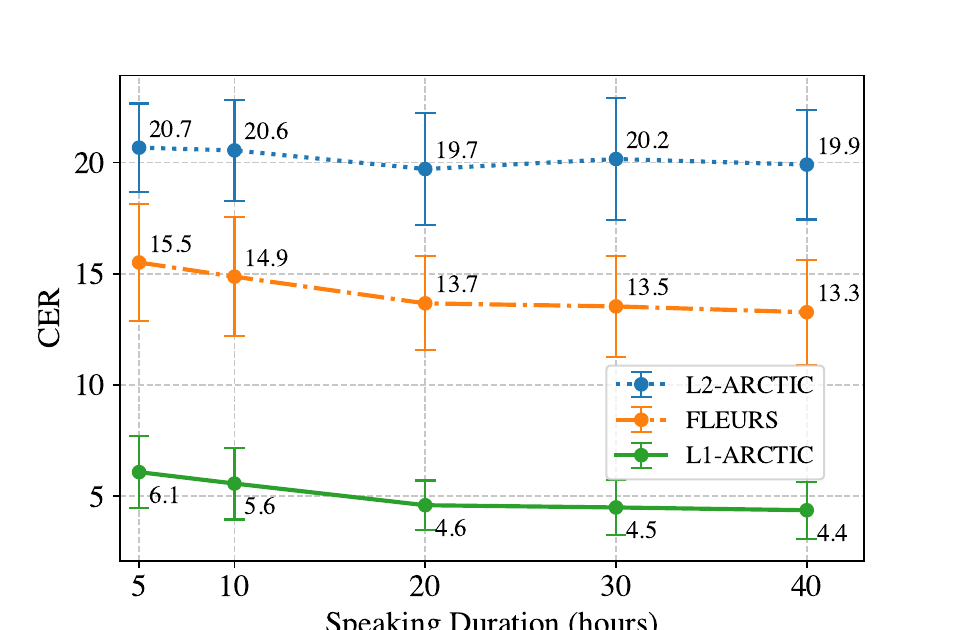}
    \caption{English CER result on three different evaluation datasets in the single-speaker training setup, where the ASR training data consist of only one speaker.}
    \label{fig:scaling-eng-hours}
\end{figure}

\begin{figure*}[!t]
    \centering
    \includegraphics[width=0.95\linewidth]{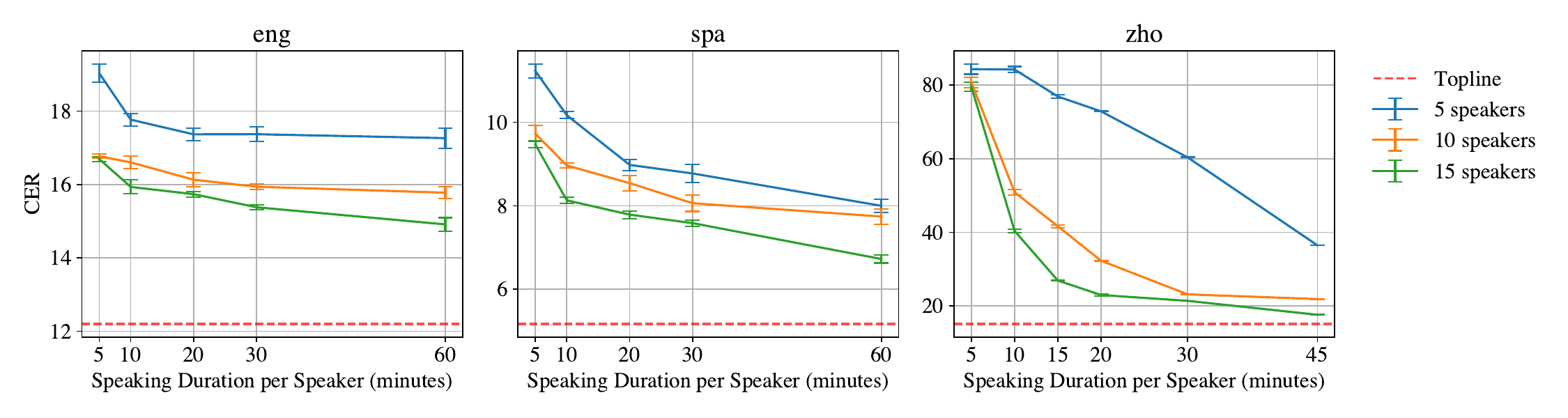}
    \caption{Effects of increasing speaker duration of each speaker on out-of-distribution accent ASR performance. 
    }
    \label{fig:more-speaking}
\end{figure*}

\begin{figure}[!t]
    \centering
    \includegraphics[width=0.99\linewidth]{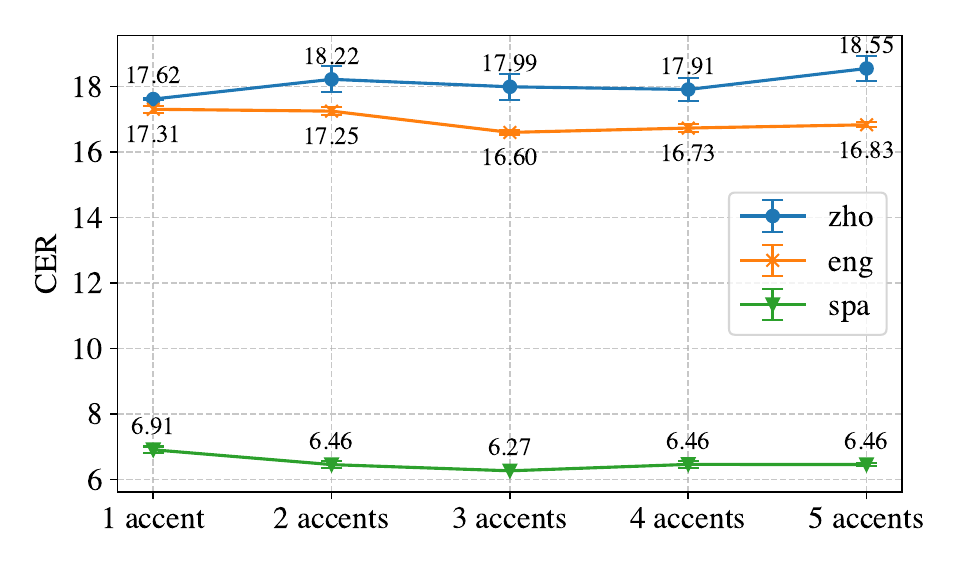}
    \caption{Effects of accent diversity, which refers to the total number of unique accents, on ASR performance for out-of-distribution accents.}
    \label{fig:accent-diversity}
\end{figure}

\section{Results and Discussion} \label{sec:results}

We find that more speakers and more training hours improve ASR performance on out-of-distribution accents. In Figure~\ref{fig:more-speakers}, we see that increasing the number of speakers while keeping the total training data size fixed consistently improves ASR performance (lower CER) on out-of-distribution accents across all three languages. For \eng, expanding from 1 to 15 speakers reduces CER from 19.8\% to 15.6\% with 5 total audio training hours. Similar improvements are observed for \spa (8.7\% to 7.4\% CER) and \zho (23.1\% to 21.3\% CER). Notably, for \eng and \spa, the performance gains are pronounced in the initial increase of speakers (for instance, 2 to 15 speakers in Spanish) before starting to plateau, suggesting that even a modest increase in speaker diversity can substantially improve ASR accent robustness.

Figure~\ref{fig:more-speakers} further demonstrates that the benefits of increased speaker count are amplified when combined with more total training hours. At higher speaker count for \eng and \spa, models trained with more hours (e.g., 20 hours for \eng, shown in green) consistently outperform those trained with fewer hours (e.g., 5 hours, shown in blue). Furthermore, for \spa, the gap between different hour conditions widens as the number of speakers increases. For \zho, doubling the training hours significantly closes the ASR performance gap with the topline.

We observe that increasing the number of speakers is preferred over increasing speaking duration per speaker when we have a fixed total training hours. Table~\ref{tab:speakers-vs-speaking} shows that when the total audio duration of the training data is kept constant, more speakers result in better ASR performance compared to more speaking per speaker. Most notably, for \zho, tripling the speaker count from 5 to 15 drops CER from 60.3 to 39.9. 

Figure~\ref{fig:scaling-eng-hours} 
demonstrates the results from single-speaker experimental setup, where we scale up the audio duration of \eng training data comprising one speaker only.
For FLEURS and L1-ARCTIC, where the test accent is in-distribution, we observe a consistent decline in CER as speaking duration increases. In contrast, for L2-ARCTIC, where the non-native accents are outside the training distribution, the CER remains at around 20\%. The results further suggest that, without sufficient speaker diversity, scaling up training data helps ASR for seen accents but not for unseen accents.

Figure~\ref{fig:more-speaking} shows that increasing the speaking duration per speaker, which leads to increased total training hours, improves ASR for out-of-distribution accents. In other words, practitioners should still aim to collect as much training data from each speaker if possible. However, Figure~\ref{fig:more-speaking} still shows that sometimes even increasing per-speaker duration may not match the performance gains from having more speakers. For instance, for \zho, having more speakers (10 speakers each speaking only for 10 minutes) have substantially lower CER compared to more per-speaker duration (5 speakers each speaking for 30 minutes) despite the former having less overall training data (100 minutes versus 150 minutes of total training data).

Surprisingly, there is minimal benefits for explicitly prioritizing accent diversity in low-resource ASR. As illustrated in Figure~\ref{fig:accent-diversity}, for \eng and \spa, the CER decreases initially and increases again when we increase coverage of unique accents (while keeping the number of speakers and training hours controlled). For \zho, increasing accent diversity does not yield any benefit to generalization to out-of-distribution accents.

\textbf{Practical Recommendations.} Our findings have important implications for data collection strategies in low-resource scenarios: (1) to handle out-of-distribution accents, increasing accent diversity is not a necessary condition; (2) prioritize recruiting more speakers over collecting longer recordings from fewer speakers to mitigate accent bias; (3) for in-distribution accents, it helps to scale up the per-speaker duration. In other words, collect as much data as possible from each speaker while maintaining speaker diversity.

\section{Conclusion}
We systematically investigated how speaker count, speaking duration per speaker, and accent diversity in training data affect ASR robustness towards out-of-distribution accents in low-resource settings. Our work provides practical recommendations on how to best allocate limited data collection resources for developing accent-robust ASR systems, particularly for under-served languages.

\bibliographystyle{IEEEtran}
\bibliography{main}

\end{document}